# Image Classification by Reinforcement Learning with Two-State Q-Learning

Abdul Mueed Hafiz[1]

[1] Department of Electronics and Communication Engineering
Institute of Technology, University of Kashmir
Srinagar, J&K, India, 190006

ORC-ID: 0000-0002-2266-3708

Email: mueedhafiz@uok.edu.in

**Abstract.** In this paper, a simple and efficient Hybrid Classifier is presented which is based on deep learning and reinforcement learning. Here, Q-Learning has been used with two states and 'two or three' actions. Other techniques found in the literature use feature map extracted from Convolutional Neural Networks and use these in the Q-states along with past history. This leads to technical difficulties in these approaches because the number of states is high due to large dimensions of the feature map. Because the proposed technique uses only two Q-states it is straightforward and consequently has much lesser number of optimization parameters, and thus also has a simple reward function. Also, the proposed technique uses novel actions for processing images as compared to other techniques found in literature. The performance of the proposed technique is compared with other recent algorithms like ResNet50, InceptionV3, etc. on popular databases including ImageNet, *Cats and Dogs* Dataset, and Caltech-101 Dataset. The proposed approach outperforms others techniques on all the datasets used.

## 1. Introduction

Reinforcement Learning (RL) [1-4] has garnered much attention [4-13]. In computer vision, good initial work [7,14,15,11,9,5,12,16-20] has been undertaken. In [14], the authors aim to reduce the large computational costs of using large images, by proposing a RL agent which adaptively selects the resolution of every image provided by the detector. They train the agent with double rewards by choosing lower resolution images for a coarse level detector in case of the image being dominated by large objects, and higher resolution images for a fine level detector in case of the image being dominated by small objects. In [20], the authors propose an object detection technique based on reinforcement Q-learning [21,22]. They use a policy search based on analytic gradient computation with continuous reward. They report almost two orders of magnitude speed-up over other popular techniques found in literature. In [23], an adaptive deep Q-learning technique has been used for improving and shortening computational time for digit recognition. They refer to their novel network as Q-learning deep belief network (Q-ADBN). This network extracts features from a deep auto-encoder [24], which are considered as current states of Q-learning algorithm. After conducting experiments on MNIST dataset [25], the authors of the above work claim that their technique is superior to other techniques in terms of accuracy and running time. In [26], the authors propose an object detection approach which uses zooming and translation for successively refining the bounding box for the objects. They use a VGG-16 Convolutional Neural Network (CNN) [27] and concatenate its feature map with a history vector, and this new map is fed to a Q-network for further processing, leading to interesting results. In [28], the authors propose a technique in which an agent learns to deform a bounding box using simple transformation actions, with the aim of obtaining specific location of objects by following top-down reasoning. The actions used are horizontal moves, vertical moves, scale changes and aspect ratio changes.

Taking a hint from the work in area of Maximum Power Point Tracking (MPPT) which is used in Photovoltaic Arrays[29], a simple and efficient technique has been proposed for image classification which gives high accuracy. It is based on deep learning as well as reinforcement learning. The technique involves using feature maps obtained from a pre-trained CNN like ResNet50 [30], InceptionV3 [31], or AlexNet [32]. Next, reinforcement learning is used for optimal action proposal generation (rotation by a specific angle or translation) on the image. After application of the final action to the original test image, and obtaining feature map from the CNN [33-36,30,31,37], classification is done using a second classifying structure, like a Support Vector Machine (SVM) [38-40] or a Neural Network (NN) which has been trained on the CNN feature maps of training images.

Many reinforcement learning techniques [20,18,26,28] used for image classification use actions like zoom and translation according to visual detection in humans. Thus they miss the important action of rotating the field of view used in human visual image comprehension. In this paper rotation of image by specific angle(s) has been used which is novel in itself. Also Q-Learning has been used in reinforcement learning. Q-states which have been used in the other approaches of reinforcement learning based object detection, use features with high dimensions combined with state history. This technique usually leads to large state-space, in turn leading to optimization problems. Addressing these problems was the motivation behind this work. The proposed technique uses only two states, and two or three actions. As a consequence of this strategy, the Q-table has two rows and two/three columns. To the best of available knowledge, this is the first technique using two Q-states, as well as using image rotation as an action. As a result, the overall task of using reinforcement learning in computer vision becomes simple and also becomes efficient. Better results are obtained in comparison to other techniques involving CNNs like ResNet50 [41,30,37], InceptionV3 [31], etc.

The major highlights of the paper are:
- A novel image recognition technique is proposed which is based on Reinforcement Learning. This is a first to the best of knowledge.
- Rotation of image has been used for image recognition, which is akin to tilting of vision. This is also a first to the best of knowledge.
- The proposed technique outperforms others on all the datasets used in the current work.

The rest of the paper is structured as follows. Section 2 discusses the proposed approach. Section 3 gives a brief description of the various datasets used in the study. Section 4 discusses the experimentation. Conclusion is presented in Section 5.

## 2. Proposed Approach

In this paper a hybrid approach of deep learning [42,7,14,43-46] and Reinforcement Learning (RL) is proposed. The CNNs used are ResNet50 [41,30,37], InceptionV3 [31], and AlexNet [32]. First, feature-map of the CNN is obtained for every training sample by feeding it to the CNN. Let the set of all of these feature maps be referred to as $F_{Train}$. A secondary classifier like a Support Vector Machine, or a Neural Network is trained on $F_{Train}$. For classifying a test sample, a filtering criterion for 'hard to classify' samples must be used. If the test sample is tagged as hard, it is classified by reinforcement learning. If not, then CNN is used for classification. This paper is not about filtering criteria, hence no such criterion has been used except that all test samples misclassified by the CNN are tagged as 'hard to classify.' Every 'hard' test sample is first fed to the CNN. Next, the feature map of the CNN, viz. $F_{Sample}$ for the test image is obtained. Reinforcement learning based classification is done as follows. A random action is selected from a bank of actions whose number does not exceed 3 in all the experiments. Next, the action permutation is applied to the test image. Then the new feature map ($F'_{Sample}$) of the permuted image is obtained from the CNN. The action permutations used in this work are mainly image rotation with specific angles and sometimes diagonal translation. For reinforcement learning, Q-Learning with random policy is used. Two states *(n = 2)* and 'two or three' actions *(a = 2) or (a = 3)* are used. The current state is decided after observing a metric viz. 'standard deviation' of the prediction scores (of the second classifier) before and after applying the permutation. Let *M* be metric for original image and $M_1$ be metric after applying current action. The new state is decided based on the criteria whether $M_1$ is lesser than, equal to, or larger than *M* respectively. The Q Table having 2 rows *(n=2)* and *a* columns *(a = 2 or 3)* is initialized to zero. Reward *r* is based on the comparison as shown below:

$$r = \begin{cases} +1, & if\ M_1 > M \\ 0, & if\ M_1 = M \\ -1, & if\ M_1 < M \end{cases} \quad (1)$$

Number of iterations for updating the Q Table is $N = a \times m$, where *a=2* or *3* and *m* is a constant (usually 20). After each iteration, the Q value entry for the current 'state-action pair' with state *s* and action *a i.e. Q(s,a)* present in the Q-Table, is updated as per the Q-Learning Update Rule:

$$Q(s,a) = Q(s,a) + \alpha[\ r + \gamma \max_{\forall b \in A} Q(s',b) - Q(s,a)\ ] \quad (2)$$

Where *s'* is new state, the learning rate $\alpha = 0.4$, and the discount rate $\gamma = 0.3$. Flowchart for the proposed RL algorithm is given in Figure 1. After completing *N* iterations of Q-Learning, optimal action is chosen as the action having highest value in Q-Table. Finally the optimal action is applied to the original sample/test image. Next the image is fed to the CNN giving its feature map. This feature map is fed to the second classifying structure for classification. Figure 2 shows the proposed approach in modular fashion.

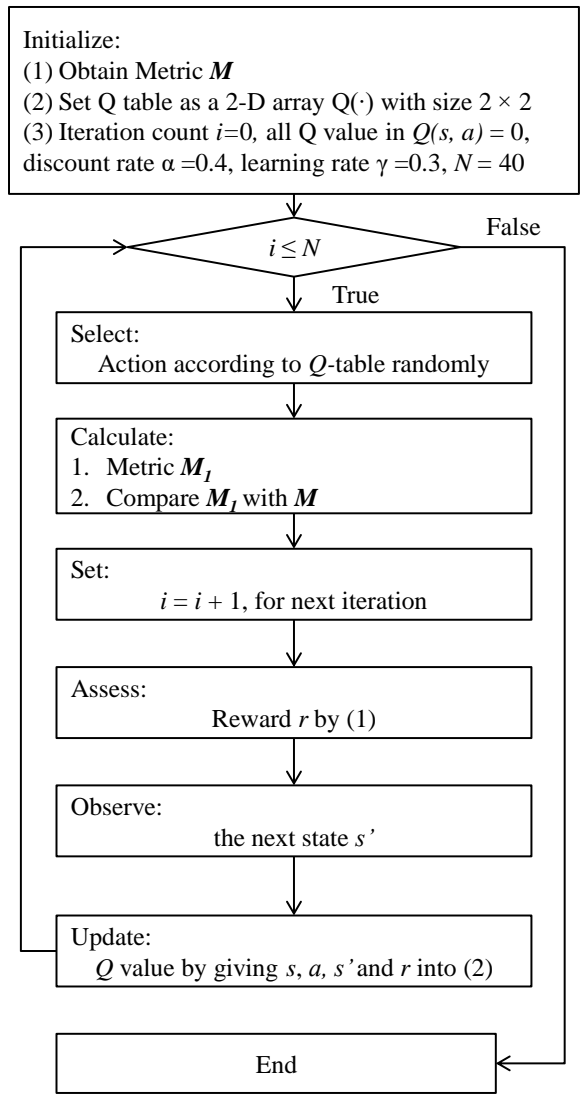

**Figure 1**. Flowchart of the proposed RL Algorithm

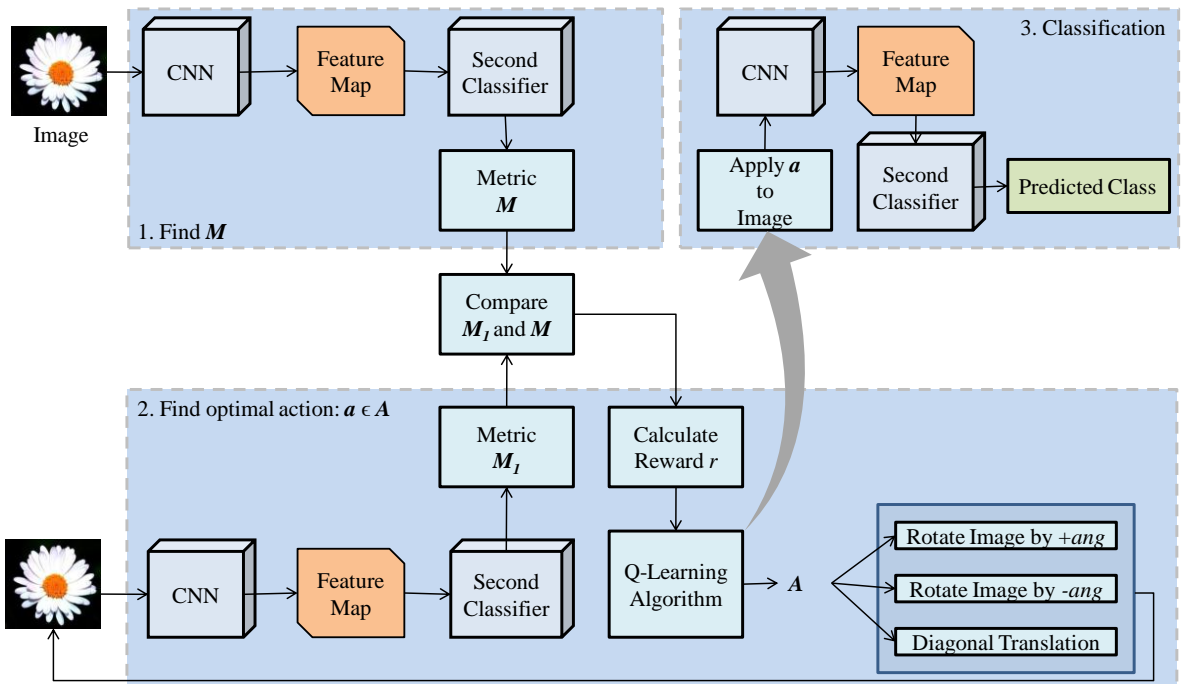

**Figure 2.** Proposed Technique

The NNs used on top of ResNet50 and InceptionV3 networks are shown in Figure 3(a) and 3(b), respectively.

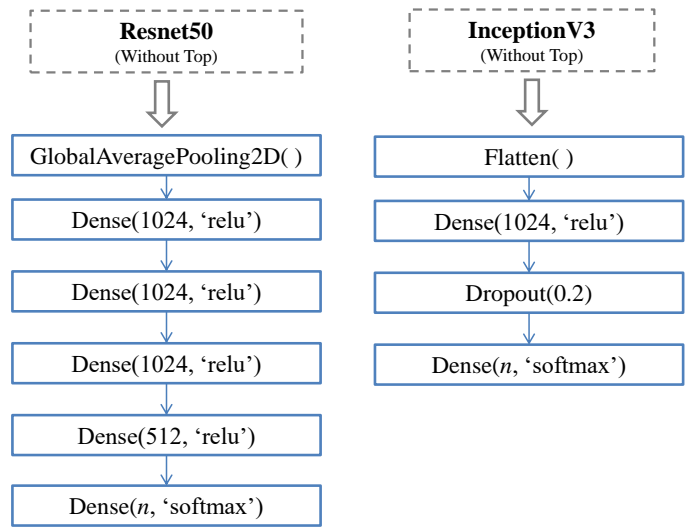

**Figure 3.** Shown are NNs used after the CNNs, viz. ResNet50 (left), and InceptionV3 (right) which are implemented in Tensorflow without top; *n* is number of classification categories

## 3. Datasets Used

Benchmarking has been done on three popular databases viz. ImageNet [47,48], *Cats and Dogs* Dataset [49], and Caltech-101 Dataset [50].

### 3.1 ImageNet

ImageNet[47] is a large image database for use in visual object recognition. In it, more than 14 million images are hand-annotated. Also at least in one million of the images, bounding boxes have also been provided. ImageNet has more than 20,000 categories with a each category typically consisting of hundreds of images. The annotation database of third-party image URLs is available directly from the ImageNet website. Since 2010, the ImageNet project runs an annual programming contest viz. the ImageNet Large Scale Visual Recognition Challenge (ILSVRC). In this various algorithms compete for correct classification and detection of objects and scenes. This challenge uses 1000 non-overlapping object categories.

### 3.2 *Cats and Dogs* Dataset

Web services have often been protected with challenges that are relatively easier for people to solve as compared to computers, e.g. CAPTCHA (Completely Automated Public Turing test to tell Computers and Humans Apart) or HIP (Human Interactive Proof). HIPs have been used for various purposes like email and blog spam reduction, and prevention of brute-force attacks on web sites having passwords. ASIRRA (Animal Species Image Recognition for Restricting Access) is a HIP that asks users to identify photographs of cats and dogs. The task is difficult for computers. However studies have shown that humans accomplish it quickly and accurately. ASIRRA has partnered with Petfinder.com, to provide Microsoft Research with more than three million images of cats and dogs, which have been manually classified by people. Kaggle offers a subset of this dataset [49] for research.

### 3.3 Caltech-101

Caltech-101 [50] is a database of images compiled by Fei-Fei Li, Marco Andreetto, Marc 'Aurelio Ranzato and Pietro Perona at the California Institute of Technology. It is intended to be used for computer vision research. It is mainly applicable to techniques for image recognition, classification and categorization. Caltech-101 contains 9,146 images, across 101 distinct object categories ( e.g. faces, watches, ants, pianos, etc.). It also has a set of annotations which describe the outlines in each image.

The distribution of data amongst the experimental setups is shown in Table 1.

**Table 1**. Distribution of Data Experimentally

| Dataset | Classes Used | Training Images | Validation Images | Testing Images |
|---|---|---|---|---|
| ImageNet [47] | 4 (Bikes, Ships, Tractors, Wagons) | 1531 | 788 | 745 |
| Cats and Dogs [49] | 2 (Cats, Dogs) | 2000 | 500 | 500 |
| Caltech-101 [51] | 50 | 750 | - | 1250 |

## 4. Experimentation

Experiments were conducted on a machine having an *Intel® Xeon®* processor (with 2 Cores), 12.6 GB RAM and 12 GB GPU. For benchmarking of the performance of the proposed technique, its performance was compared with that of the pre-trained CNN used alone after being fine-tuned on the datasets. Tensorflow [52] has been used for implementing the CNNs (pre-trained, having ImageNet weights) and algorithms. For training the CNNs using transfer learning, 10 *training epochs* were used, with optimizer: *Adam*, Loss: *Categorical Crossentropy*, and Learning Rate: *0.001*. Benchmarking has been done on three popular databases viz. ImageNet [47,48], *Cats and Dogs* Dataset [49], and Caltech-101 Dataset [50].

Tables 2 to 4 show the benchmarking for the performance of the proposed approach on the datasets used. It should be noted that for conventional CNN usage no rotation is done during evaluation. It should be noted that for **Caltech-101**, a two action set comprising of angular rotation by 12.5° or by -12.5° gave best results. For **ImageNet** and *Cats and Dogs* **Dataset**, a three action set comprising of angular rotation by 90°, or by 180°, or *downward and rightward* diagonal translation by 15 pixels gave best results. The action sets gave best results, as compared to others including no-rotation action, etc. It should be noted that rotation permutation during evaluation leads to better results than the conventional CNNs which do not use rotation. This can be due to alignment between test image and previous training image (due to rotation during evaluation) in the CNN. One advantage of this technique is that training need not be done extensively, thus saving resources like memory and time considerably for systems having both offline and online training. This statement is subject to speculation and will be more revealed and explained in future work.

**Table 2**. Classification Accuracy of Various Approaches on ImageNet
(**Second Classifier Used**: NN; **Metric Used**: Std. Deviation of Softmax scores, **Feature Map Size**: 1x1024)

| Approach | Secondary NN used on top of Layer | Image Size | Accuracy |
|---|---|---|---|
| ResNet50 | #174: @(conv5_block3_out) | 150x150x3 | .8242 |
| Proposed Approach using ResNet50 | #174: @(conv5_block3_out) | 150x150x3 | **.8309** |
| Inception V3 | #228: @(mixed7) | 150x150x3 | .8564 |
| Proposed Approach using InceptionV3 | #228: @(mixed7) | 150x150x3 | **.8644** |

**Table 3**. Classification Accuracy of Various Approaches on *Cats and Dogs* Dataset (**Second Classifier Used**: NN; **Metric Used**: Std. Deviation of Softmax scores, **Feature Map Size:** 1x1024)

| Approach | Secondary NN used on top of Layer | Image Size | Accuracy |
|---|---|---|---|
| ResNet50 | #174: @(conv5_block3_out) | 224x224x3 | .9780 |
| Proposed Approach using ResNet50 | #174: @(conv5_block3_out) | 224x224x3 | **.9860** |
| Inception V3 | #228: @(mixed7) | 150x150x3 | .9440 |
| Proposed Approach using InceptionV3 | #228: @(mixed7) | 150x150x3 | **.9520** |

As is observed from Table 2, larger image sizes lead to higher accuracies. However, the training times also increase.

**Table 4**. Classification Accuracy of Various Approaches on Caltech-101 Dataset (**Second Classifier Used**: Binary-SVM Ensemble; **Metric Used**: Std. Deviation of SVM prediction scores, **Feature Map Size:** 1x4096)

| Approach | Secondary SVM used on top of Layer | Image Size | Accuracy |
|---|---|---|---|
| AlexNet | - | 227x227x3 | .841 |
| CNN-SVM Hybrid Approach [53] | #20: @(fc7) | 227x227x3 | .882 |
| Proposed Approach using AlexNet | #20: @(fc7) | 227x227x3 | **.898** |

As is observable from Table 2 to 4, the proposed approach outperforms other approaches on all the datasets used. This technique can pave the way for a novel image recognition approach using tilt of vision in classifiers, etc. Also, the use of reinforcement learning in computer vision is a first, to the best of available knowledge. This technique can pave the way for a whole new generation of computer vision classifiers based on reinforcement learning. It should also be noted that in this study dimensional reduction [54] is not used. Also, that the proposed approach is instance-based. Thus the processing time is more as compared to other techniques in this area. This is one of the limitations which is intended to be addressed in future work. In future, more work will be done on using larger datasets. Also, work would be done on making the proposed approach faster by using techniques like dimensional reduction, or using smaller feature maps. Also, work would be done on using the proposed approach on other interesting computer vision tasks like instance segmentation [55], etc.

## 5. Conclusion and Future Work

In this paper, a straightforward and efficient learning system is investigated which combines deep learning with reinforcement learning. The proposed technique is simpler than other contemporary techniques found elsewhere. This is for the reason that others use high number of states while as the proposed approach uses only two states. Thus optimization is easy and the reward function is straightforward. Other approaches use visualization tasks like zoom and translation. A novel technique i.e. rotation has been used which is similar to tilt of visual field. Three databases have been used in the experimentation here. These are ImageNet, *Cats and Dogs* Dataset, and Caltech-101 Dataset. Benchmarking of the proposed classifier has been done. The proposed approach outperforms other approaches including ResNet50, InceptionV3, etc. on all the three datasets used.

## Declarations

The authors declare no conflict of interest.